\documentclass[letterpaper]{article} 
\usepackage{aaai2027}  
\usepackage[hyphens]{url}  
\usepackage{graphicx} 
\usepackage{natbib}  
\usepackage{caption} 
\usepackage{algorithm}
\usepackage{algorithmic}
\usepackage{amsfonts}
\usepackage{amsmath}
\usepackage{multirow}
\usepackage{tabularray}
\usepackage{svg}
\usepackage{subcaption}
\usepackage{rotating}
\usepackage[table]{xcolor}

\usepackage{newfloat}
\usepackage{listings}
\DeclareCaptionStyle{ruled}{labelfont=normalfont,labelsep=colon,strut=off} 
\floatstyle{ruled}
\newfloat{listing}{tb}{lst}{}
\floatname{listing}{Listing}

\usepackage{booktabs}

\title{When Point Clouds Outperform Pixels: Rethinking Zero-Shot Multimodal Anomaly Detection}
\author {
    Chenglin Ye\textsuperscript{\rm 1}\equalcontrib,
    Lupeng Liu\textsuperscript{\rm 1}\equalcontrib,
    Dongbo Yu\textsuperscript{\rm 1},
    Jun Xiao\textsuperscript{\rm 1}\corresponding,
    Yunbiao Wang\textsuperscript{\rm 1}\corresponding
}
\affiliations {
    \textsuperscript{\rm 1}School of Artificial Intelligence, University of Chinese Academy of Sciences\\
    yechenglin24@mails.ucas.ac.cn, \{liulupeng, yudongbo, xiaojun, wangyunbiao\}@ucas.ac.cn
}

\begin{document}

\maketitle

\begin{abstract}
Zero-shot multimodal anomaly detection commonly assumes that RGB and point cloud modalities are equally reliable and can contribute uniformly to anomaly localization. We challenge this assumption. Using a set of recently proposed stringent metrics that penalize false anomaly responses in normal regions, we find that point clouds are substantially more reliable than RGB under zero-shot category shift. Motivated by this observation, we propose WOOPS (\textbf{W}hen P\textbf{o}int Cl\textbf{o}uds Out\textbf{p}erform Pixel\textbf{s}), a reliability-aware zero-shot multimodal anomaly detection framework. To strengthen the more reliable geometric modality, we design a Multi-view Information Decoupling module to suppress heterogeneous information from multi-view point cloud projections and enhance point cloud feature quality. To avoid unconditional fusion, we further introduce a Modality Reliability Calibration module to adaptively calibrate modality contributions according to their reliability. Extensive experiments show that our method achieves the best or competitive performance under the new metrics in both unimodal and multimodal settings. Further analysis demonstrates that point cloud information also improves RGB-only inference, while ablations verify the effectiveness of both modules. Code will be released upon acceptance.
\end{abstract}


\begin{figure}[ht]
    \centering
    \begin{subfigure}[t]{\linewidth}
        \centering
        \includegraphics[width=0.9\linewidth]{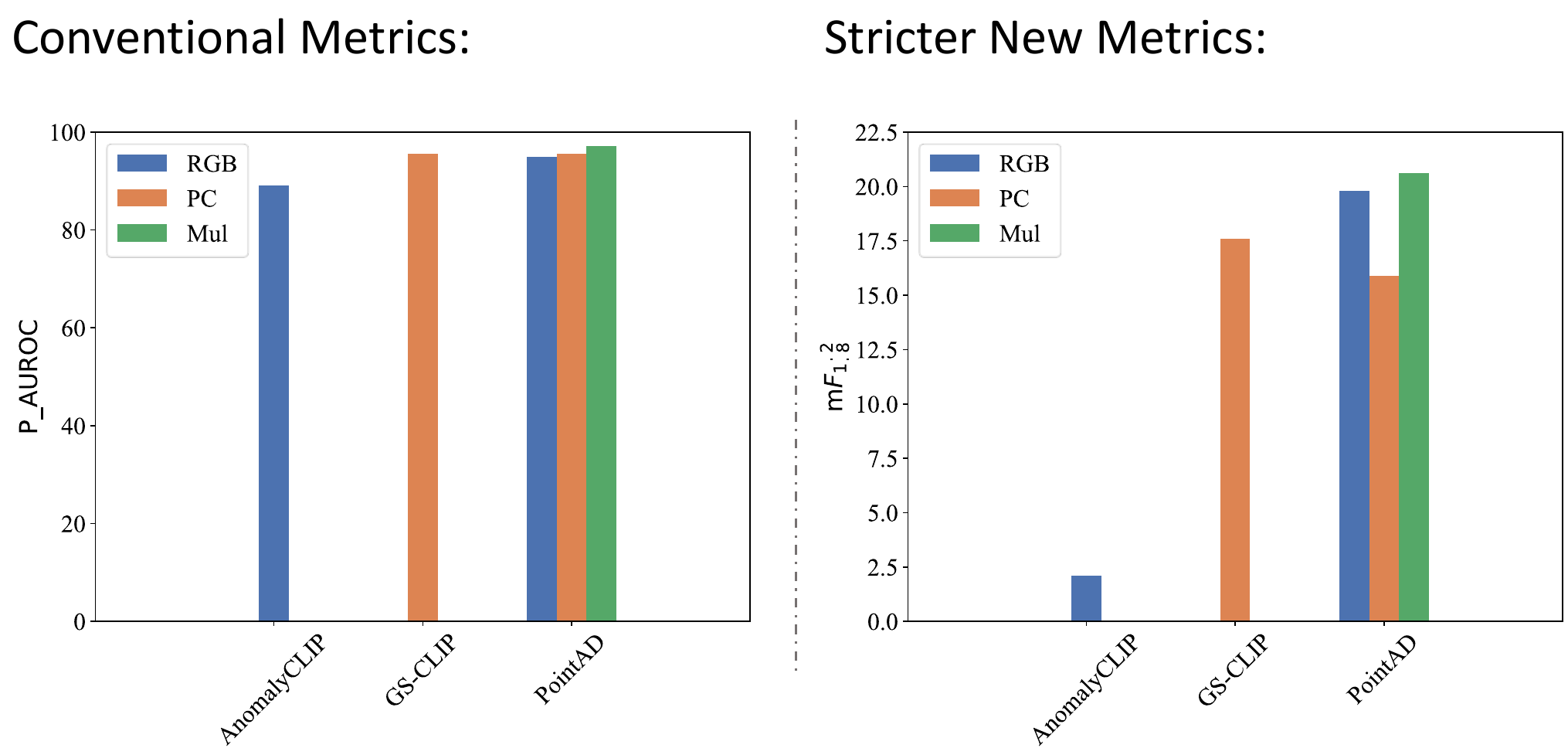}  
        \caption{Revisiting Zero-Shot AD}
    \end{subfigure}
    
    \vspace{0.05cm}  
    
    \begin{subfigure}[b]{\linewidth}
        \centering
        \includegraphics[width=0.9\linewidth]{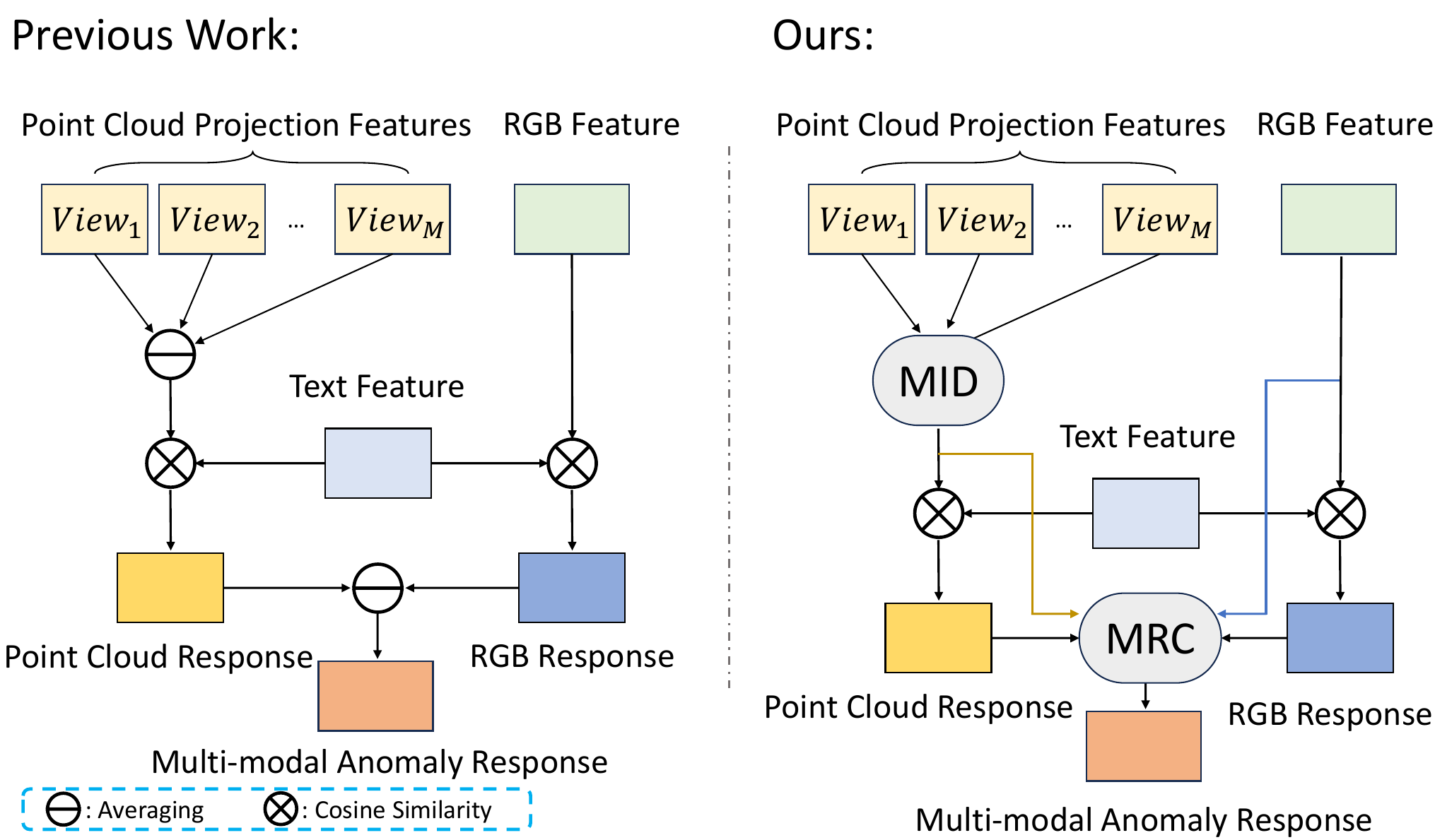}  
        \caption{Architectural Comparison}
    \end{subfigure}
    
    \caption{(a) We illustrate results using P-AUROC as a representative of conventional metrics\cite{m3dad} and ${\mathrm{m}F_\mathrm{1}}^{.2}_{.8}$ as a representative of the new evaluation protocol\cite{metric}. Please zoom in for details. (b) Comparison of network architectures between our model and existing zero-shot multimodal anomaly detection methods.}
    \label{fig:intro}
\end{figure}

\section{Introduction}

Anomaly Detection (AD) aims to identify and localize anomalous characteristics or defects in target objects, playing a pivotal role in applications such as industrial quality inspection\cite{10.3389/frobt.2025.1554196, faprompt, 11092978, Fang_Su_Lv_Xu_Yu_2025} and medical diagnostics\cite{10376801, CAI2025103500, Gu_Zhu_Zhu_Chen_Ge_Tang_Wang_2026}. Zero-shot AD further considers a more practical setting where test categories are unseen during training, requiring models to learn transferable notions of normality across diverse objects. Compared with conventional settings, this imposes higher demands on representation generalization, modality robustness, and response calibration.

RGB and point cloud data provide complementary information: RGB captures texture and appearance, while point clouds encode geometric structure. Recent zero-shot multimodal methods leverage both by projecting point clouds into multiple views and extracting features via image encoders\cite{pointad, zuma, gs-clip}, often with vision-language models such as CLIP\cite{clip}. However, multi-view projections introduce heterogeneous information arising from variations in rendering viewpoints, such as shape discrepancies caused by self-occlusion or viewpoint divergence. Direct aggregation of such views can degrade point cloud representations and weaken geometric anomaly responses, especially under zero-shot category shifts where robust geometry is critical. Another issue is that the reliability of each modality in zero-shot AD remains unclear. Existing methods often assume that combining modalities consistently improves performance, but this may not hold in zero-shot settings. RGB responses are sensitive to appearance variations and can produce false positives in normal regions\cite{m3dad}. Without modeling modality reliability, fusing such responses with point cloud outputs may amplify errors.

In this work, we first revisit zero-shot AD from the perspective of modality reliability under a stricter evaluation protocol \cite{metric}. Different from conventional metrics\cite{m3dad} that mainly focus on anomalous areas, the newly introduced evaluation criterion places stronger emphasis on suppressing anomaly responses in normal regions. As presented in Fig. \ref{fig:intro}, we discover that point cloud-only inference can significantly outperform RGB-only inference, suggesting that geometry plays a dominant role in cross-category generalization, while RGB may introduce unreliable responses. Moreover, training with point cloud data improves performance even for RGB-only inference, highlighting its contribution to transferable representations.

Considering the existing issues and the above experimental analysis, we propose WOOPS (\textbf{W}hen P\textbf{o}int Cl\textbf{o}uds Out\textbf{p}erform Pixel\textbf{s}), a geometry-centric and reliability-aware zero-shot multimodal AD framework. Specifically, we propose a Multi-view Information Decoupling (MID) module to alleviate the interference caused by heterogeneous information across different views, thereby enhancing the quality of point cloud features. Furthermore, we design a Modality Reliability Calibration (MRC) module that dynamically calibrates modality contributions according to their reliability, suppressing unstable activations and improving cross-domain robustness.

Extensive experiments on widely adopted benchmarks demonstrate that our method achieves the best or competitive performance under the stricter anomaly evaluation metrics. Moreover, comprehensive analyses show that point cloud data benefits RGB inference, MID improves detection performance, and MRC enhances robustness. Although a trade-off exists under conventional metrics, our method significantly reduces false positives under stricter evaluation, which is crucial for practical anomaly localization.

Our contributions are summarized as follows:
\begin{itemize}
\item We revisit zero-shot multimodal AD under a stricter evaluation protocol and reveal that modality reliability is highly imbalanced under zero-shot category shift, where point cloud representations provide more transferable anomaly cues than RGB features.
\item We propose the Multi-view Information Decoupling (MID) module, a point cloud feature enhancement module that mitigates heterogeneous information from multi-view projections and improves geometric anomaly representation.
\item We develop the Modality Reliability Calibration (MRC) module, a modality response balancing module that explicitly models modality reliability and improves robustness across datasets and modality settings.
\item Extensive experiments under a newly proposed stringent evaluation protocol show that WOOPS achieves state-of-the-art performance in both unimodal and multimodal settings on two widely adopted multimodal AD benchmarks.
\end{itemize}

\section{Related Work}
\subsection{Multimodal Anomaly Detection}

Multimodal anomaly detection typically follows unsupervised or self-supervised paradigms \cite{10.1007/978-3-031-72627-9_5, 11091585, LI2025131243} and can be broadly grouped into three categories. The first line is feature-embedding-based methods \cite{Horwitz_2023_CVPR, Gu_Zhang_Liu_Chen_Peng_Gan_Jiang_Shu_Wang_Ma_2024, Rudolph_2023_WACV}, which characterize the distribution or decision boundary of normal samples and treat deviations therefrom as anomalies. For instance, LPFSTNet\cite{CHENG2025129408} and AST\cite{Rudolph_2023_WACV} employ teacher–student networks to model the normal distribution, while M3DM\cite{Wang_2023_CVPR} and M3DM-NR\cite{11091585} delimit normality with a memory bank. The second line is reconstruction-based methods\cite{ZHAO_2025_CVPR, Zavrtanik_2024_WACV, LIU2025126665, cfm}, which leverage the prior that a model trained exclusively on normal data fails to reconstruct anomalous regions, and thus localize anomalies via reconstruction error. Representative works such as CFM\cite{cfm} and MODMAP\cite{Costanzino_2026_CVPR} reconstruct features through cross-modal mapping and identify anomalies from the discrepancy between reconstructed and extracted features. The third line consists of discriminative methods that perform self-supervision by synthesizing pseudo-anomalous samples\cite{ASAD2025103284, 10.1007/978-3-031-72627-9_5, WANG2025130073, 10.1145/3581783.3611876, 10.1145/3746027.3755261}. Nevertheless, these approaches struggle to generalize to unseen objects, rendering them less applicable under privacy-sensitive (e.g., industrial confidentiality) or data-scarce (e.g., brand-new products) scenarios.


\subsection{Zero-Shot Anomaly Detection}

Zero-shot anomaly detection exploits the open-world knowledge encapsulated in vision–language models\cite{clip, ulip, ulip2} to generalize to unseen objects. While zero-shot 2D anomaly detection has matured considerably\cite{anomalyclip, adaclip, winclip, Hou_2026_CVPR, Chen_2026_CVPR, faprompt}, zero-shot multimodal anomaly detection remains underexplored, with two notable gaps. The first concerns point cloud feature extraction. Since CLIP cannot directly consume point clouds, a common workaround is to render the point cloud into multi-view images\cite{pointad, zuma, gs-clip} and treat the multi-view image features as a surrogate for point-cloud representation. However, varying rendering views induce inconsistent information across views, degrading the quality of the resulting point-cloud features. Although BTP\cite{Li_2026_CVPR} and MuSc-V2\cite{11498451} explore large models natively supporting point-cloud modality, their deployment overhead limits practical applicability. The second gap lies in the lack of modality reliability modeling. Existing zero-shot multimodal methods predominantly assume equal reliability across all modalities \cite{pointad, zuma}, aggregating multimodal responses via simple averaging. This assumption undermines robustness when anomalies are only discernible from a specific modality.

\begin{figure*}[t]
\centering
\includegraphics[width=0.9\textwidth]{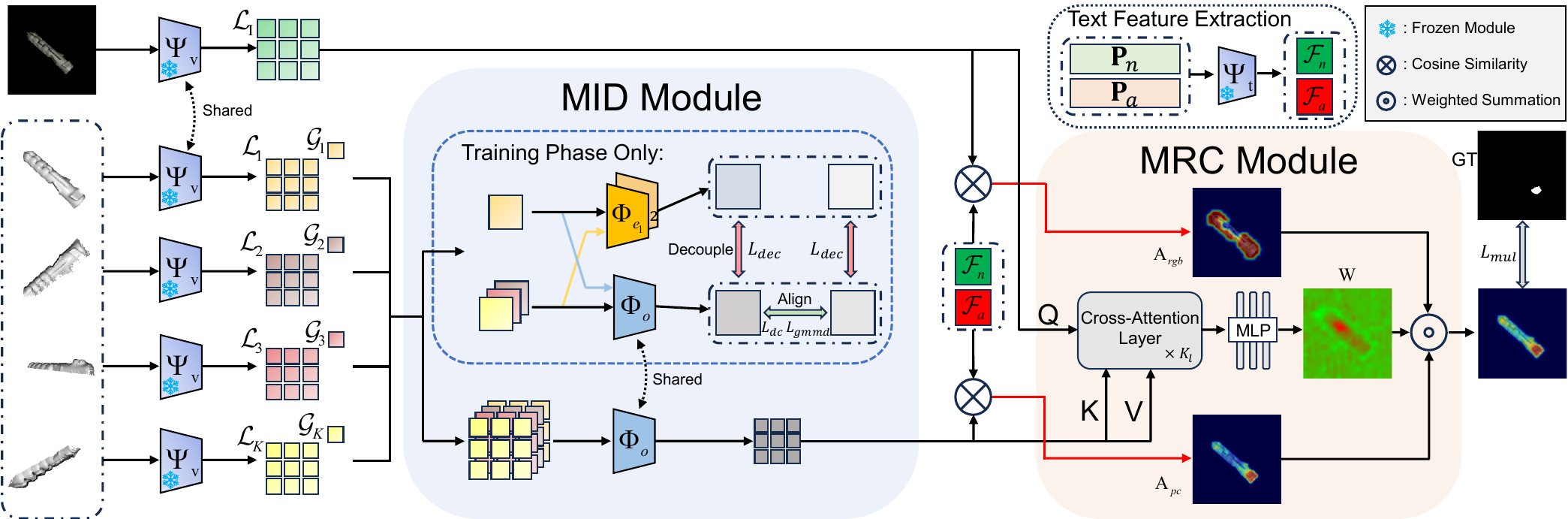} 
\caption{The pipeline of WOOPS. WOOPS employs a frozen CLIP vision encoder $\Psi_\mathrm{v}$ to extract RGB and point cloud features from images and multi-view projections, respectively, while utilizing a frozen CLIP text encoder $\Psi_\mathrm{t}$ to derive text features from learnable normal and abnormal language vectors ($\mathbf{P}_n$ and $\mathbf{P}_a$). Anomaly responses are computed via vision-language similarity. Furthermore, we propose a Multi-view Information Decoupling (MID) module to enhance point cloud features and a Modality Reliability Calibration (MRC) module to effectively calibrate anomaly responses across modalities.}
\label{fig:pipeline}
\end{figure*}

\section{Methodology}

\subsection{Overview}

We investigate zero-shot multimodal AD utilizing RGB images and point clouds.

Let $\mathfrak{D} _{tr}=\{\mathrm{O}^{tr}_i=(\mathrm{I}_i,\mathrm{P}_i,\mathrm{I}^{gt}_i,\mathrm{P}^{gt}_i)\}^{N_{tr}}_{i=1}$ and $\mathfrak{D} _{te}=\{\mathrm{O}^{te}_i=(\mathrm{I}_i,\mathrm{P}_i,\mathrm{I}^{gt}_i,\mathrm{P}^{gt}_i)\}^{N_{te}}_{i=1}$ denote the training and test sets, respectively, where $\mathrm{I} \in \mathbb{R}^{H \times W}$, $\mathrm{P} \in \mathbb{R}^{N_{pc} \times 3}$, $\mathrm{I}^{gt} \in \mathbb{R}^{H \times W}$, and $\mathrm{P}^{gt} \in \mathbb{R}^{N_{pc} \times 1}$ correspond to the RGB image, point cloud, RGB ground truth, and point cloud ground truth. Crucially, the object categories in the training and test sets are entirely disjoint.


Given a test sample, the goal is to produce an anomaly response map $\mathrm{S}$ that assigns high responses to abnormal regions and suppresses responses in anomaly-free regions.

Our framework, denoted as WOOPS (\textbf{W}hen P\textbf{o}int Cl\textbf{o}uds Out\textbf{p}erform Pixel\textbf{s}), contains four core components: Input Feature Extraction, Heterogeneity-Aware Point Cloud Feature Enhancement, Anomaly Response and Reliability-Aware Modality Response Balancing. The overall pipeline is shown in Fig. \ref{fig:pipeline}. Details about each component are as follows.


\subsection{Input Feature Extraction}

To extract point cloud features, we adopt the prevalent multi-view projection paradigm commonly used in zero-shot multimodal methods\cite{pointad, zuma, gs-clip}. Specifically, given the point cloud input $\mathrm{P}$ and its corresponding ground truth $\mathrm{P}^{gt}$ for the training sample $(\mathrm{I},\mathrm{P},\mathrm{I}^{gt},\mathrm{P}^{gt}) \in \mathfrak{D} _{tr}$, we first render them into $K$ projected views: 
\begin{equation}
\mathrm{R}_k=\Pi_k(\mathrm{P}), \mathrm{Y}_k=\Pi_k(\mathrm{P}^{gt}), k=1,\ldots,K,
\end{equation}
where $\Pi_k(\cdot)$ denotes the rendering matrix for the $k$-th view, while $\mathrm{R}_k \in \mathbb{R} ^{H\times W}$ and $\mathrm{Y}_k \in \mathbb{R} ^{H\times W}$ represent the resulting 2D rendered image and its corresponding pixel-level ground truth, respectively. For detailed information, please refer to PointAD\cite{pointad}.


Subsequently, features of $\mathrm{I}$ and $\mathrm{R}_k$ are extracted using the visual encoder $\Psi_\mathrm{v}$ of CLIP\cite{clip} to serve as inputs for the subsequent network. This process is formulated as follows:
\begin{equation}
    \mathcal{G}_\mathrm{I},\{\mathcal{L}^{(m)}_\mathrm{I}\}^M_{m=1}=\Psi_\mathrm{v}(\mathrm{I}),
\end{equation}
\begin{equation}
    \mathcal{G}_k,\{\mathcal{L}^{(m)}_k\}^M_{m=1}=\Psi_\mathrm{v}(\mathrm{R}_k), \quad k=1,2,\dots,K,
\end{equation}
where $\mathrm{I}$ is the RGB image input; $\mathcal{G}_\mathrm{I} \in \mathbb{R} ^{1 \times D}$ and $\{\mathcal{L}^{(m)}_\mathrm{I}\} \subset \mathbb{R} ^{1 \times D}$ are the global feature and local patch features of I, respectively; $\mathcal{G}_k \in \mathbb{R} ^{1 \times D}$ and $\{\mathcal{L}^{(m)}_k\} \subset \mathbb{R} ^{1 \times D}$ are those of $\mathrm{R}_k$; $D$ represents the feature dimension; and $\Psi_\mathrm{v}$ denotes the frozen CLIP visual encoder.


\subsection{Heterogeneity-Aware Point Cloud Feature Enhancement}

Beyond discriminative geometric structures, each projected view inherently contains heterogeneous information arising from variations in rendering viewpoints. To address this issue, we propose Multi-view Information Decoupling (MID) module, a heterogeneity-aware point cloud feature enhancement module. The objective of MID module is to suppress unreliable view-specific information and emphasize geometrically consistent cues across projections. MID module operates between randomly sampled views and the aggregated statistics of all other views. The core idea is to decompose each view’s local features into two complementary subspaces:
\textbf{(i)} a homogeneous subspace, which encodes shared geometric structures preserved across viewpoints, and
\textbf{(ii)} a heterogeneous subspace, which captures view-specific irrelevant information.


Specifically, we first randomly sample $X$ views from the $K$ available projection views, denoted as $\mathcal{R}=\{R_{x_1}, R_{x_2}, \dots, R_{x_X}\}$. The average global feature $\bar{\mathcal{G}}$ and average local patch features $\{\bar{\mathcal{L}}^{(m)}\}$ of these sampled views are calculated as follows:
\begin{equation}
    \mathcal{\bar{G} }= \frac{1}{X} \sum_{R_i \in \mathcal{R} }^{} \mathcal{G}_i,
\end{equation}
\begin{equation}
    \bar{\mathcal{L}}^{(m)}=\frac{1}{X} \sum_{R_i \in \mathcal{R} }^{} \mathcal{L}_i^{(m)}, \quad m=1, 2, \dots, M.
\end{equation}
Similarly, the average global feature $\bar{\mathcal{G}_A}$ and average local patch features $\bar{\mathcal{L}_A}^{(m)}$ for the remaining views are obtained using the same procedure. To extract homogeneous and heterogeneous features from both the sampled subset and the remaining views, we employ a shared homogeneous subspace encoder $\Phi _o$ along with two independent heterogeneous subspace encoders $\Phi _{e_1}$ and $\Phi _{e_2}$. All three encoders share an identical architecture, implemented as two-layer perceptrons activated by the ReLU function, with a hidden dimension of $D$. This process is formulated as follows:
\begin{equation}
\begin{split}
    \mathcal{H}^o_{G}=\Phi _o(\bar{\mathcal{G}}), &\quad \mathcal{H}^e_{G}=\Phi _{e_1}(\bar{\mathcal{G}}),\\
    \mathcal{H}^o_{AG}=\Phi _o(\bar{\mathcal{G}_A}), &\quad \mathcal{H}^e_{AG}= \Phi _{e_2}(\bar{\mathcal{G}_A}), \\
    \mathcal{H}^{o(m)}_{L}=\Phi _o(\bar{\mathcal{L}}^{(m)}), &\quad \mathcal{H}^{e(m)}_{L}= \Phi _{e_1}(\bar{\mathcal{L}}^{(m)}), \\
    \mathcal{H}^{o(m)}_{AL}=\Phi _o(\bar{\mathcal{L}_A}^{(m)}), &\quad \mathcal{H}^{e(m)}_{AL}= \Phi _{e_2}(\bar{\mathcal{L}_A}^{(m)}).
\end{split}
\end{equation}
Here, $m=1, 2, \dots, M$; $\mathcal{H}^o_{G}$ and $\mathcal{H}^e_{G}$ denote the global homogeneous feature and global heterogeneous feature of the sampled views, respectively; $\mathcal{H}^{o(m)}_{L}$ and $\mathcal{H}^{e(m)}_{L}$ represent the local patch homogeneous features and local patch heterogeneous features of the sampled views, respectively. Correspondingly, $\mathcal{H}^o_{AG}$ and $\mathcal{H}^e_{AG}$ denote the global homogeneous and heterogeneous features of the remaining views, while $\mathcal{H}^{o(m)}_{AL}$ and $\mathcal{H}^{e(m)}_{AL}$ refer to the local patch homogeneous and heterogeneous features of the remaining views.


To ensure sufficient disentanglement between homogeneous and heterogeneous features, we design a decoupling loss $L_{dec}$. To avoid prohibitive computational costs, instead of modeling distributions or calculating mutual information, we employ simple yet effective cosine similarity to quantify their potential overlap. The decoupling process is formulated as follows:
\begin{equation}
\begin{split}
    L_{dec}&=\frac{\mathcal{H}^o_{G} \cdot {\mathcal{H}^e_{G}}^\top}{\|\mathcal{H}^o_{G}\| \cdot \|\mathcal{H}^e_{G}\|} +\frac{\mathcal{H}^o_{AG} \cdot {\mathcal{H}^e_{AG}}^\top}{\|\mathcal{H}^o_{AG}\| \cdot \|\mathcal{H}^e_{AG}\|} \\
    &+\sum_{m=1}^{M} \frac{\mathcal{H}^{o(m)}_{L} \cdot {\mathcal{H}^{e(m)}_{L}}^\top}{\|\mathcal{H}^{o(m)}_{L}\| \cdot \|\mathcal{H}^{e(m)}_{L}\|} \\
    &+\sum_{m=1}^{M} \frac{\mathcal{H}^{o(m)}_{AL} \cdot {\mathcal{H}^{e(m)}_{AL}}^\top}{\|\mathcal{H}^{o(m)}_{AL}\| \cdot \|\mathcal{H}^{e(m)}_{AL}\|}.
\end{split}
\end{equation}

Inspired by DecAlign\cite{qian2026decalign}, we leverage Gaussian distributions to model the feature spaces, aiming to facilitate the alignment between the global homogeneous features $\mathcal{H}^o_{G}$ and $\mathcal{H}^o_{AG}$. This strategy is designed to strengthen the encoder $\Phi _o$'s ability to capture geometrically consistent information. Concretely, we approximate the distributions of $\mathcal{H}^o_{G}$ and $\mathcal{H}^o_{AG}$ as $\mathcal{N}_1(\mu_1,\Sigma_1)$ and $\mathcal{N}_2(\mu_2,\Sigma_2)$, respectively, with $\mu$ and $\Sigma$ representing the mean and variance. The calculation formulas for $\mu$ and $\Sigma$ are provided in the Supplement.
To enforce similarity between these two distributions, we employ a distribution consistency loss $L_{dc}$ and a Gaussian kernel-based Maximum Mean Discrepancy (GMMD) loss $L_{gmmd}$. These are defined as follows:
\begin{equation}
    L_{dc}=\|\mu_1-\mu_2\|^2+\|\Sigma_1-\Sigma_2\|^2_F,
\end{equation}
\begin{equation}
\begin{split}
    L_{gmmd}&=\mathbb{E}_{\mathbf{f},\mathbf{f}^{'}\sim \mathcal{N}_1}[k(\mathbf{f},\mathbf{f}^{'})]+\mathbb{E}_{\mathbf{g},\mathbf{g}^{'}\sim \mathcal{N}_2}[k(\mathbf{g},\mathbf{g}^{'})] \\
    &-2\mathbb{E}_{\mathbf{f}\sim \mathcal{N}_1,\mathbf{g}\sim \mathcal{N}_2}[k(\mathbf{f},\mathbf{g})].
\end{split}
\end{equation}
The aforementioned three loss functions collectively constitute the point cloud feature enhancement loss $L_{fe}$:
\begin{equation}
    L_{fe}=L_{dec}+L_{dc}+L_{gmmd}.
\end{equation}
For simplicity, we do not introduce individual weighting hyperparameters for each loss term.

Ultimately, the homogeneous subspace encoder is employed to decouple the features derived from the multi-view projections of the point cloud. The resulting homogeneous features $\hat{\mathcal{G}}_k$ and $\{\hat{\mathcal{L}}^{(m)}_k\}^M_{m=1}$ correspond to the enhanced global point cloud feature and the enhanced local patch feature, respectively, which are formulated as follows:
\begin{equation}
    \hat{\mathcal{G}}_k=\Phi _o(\mathcal{G}_k), \quad \hat{\mathcal{L}}^{(m)}_k=\Phi _o(\mathcal{L}^{(m)}_k),
\end{equation}
where $k=1,2,\dots,K$ and $m=1,2,\dots,M$.

\subsection{Anomaly Response Map}

Inspired by prior works \cite{Zhou_2022, anomalyclip}, we introduce two semantic-complementary learnable language vectors $\mathbf{P}_n$ and $\mathbf{P}_a$ of length $l$ to store generalizable normal and anomalous semantics. To prevent overfitting to the training categories, we adopt a category-agnostic initialization strategy\cite{pointad}. Using the CLIP text encoder $\Psi_\mathrm{t}$, we extract the corresponding text features $\mathcal{F}_n$ and $\mathcal{F}_a$. This process is formulated as:

\begin{equation}
    \mathcal{F}_n=\Psi_\mathrm{t}(\mathbf{P}_n), \quad \mathcal{F}_a=\Psi_\mathrm{t}(\mathbf{P}_a)
\end{equation}


The anomaly response map $\mathrm{S}$ is generated by computing the similarity between local patch features and text features for both image and point cloud modalities. For the RGB input, we first calculate the cosine similarity between $\{\mathcal{L}^{(m)}_\mathrm{I}\}$ and $\mathcal{F}_n,\mathcal{F}_a$, followed by Softmax normalization and upsampling to obtain the anomaly score map $\mathrm{A}_{rgb} \in \mathbb{R}^{H \times W}$ and the anomaly response map $\mathrm{S}_{rgb} \in \mathbb{R}^{H \times W}$, formulated as:
\begin{equation}
    \mathrm{A}_{rgb}=\frac{\operatorname{exp}(<\{\mathcal{L}^{(m)}_\mathrm{I}\},\mathcal{F}_a >/\tau ) }{\sum_{i\in {\{n,a\}}}\operatorname{exp}(<\{\mathcal{L}^{(m)}_\mathrm{I}\},\mathcal{F}_i >/\tau)}
\end{equation}
\begin{equation}
    \mathrm{S}_{rgb}= \operatorname{Bilinear}[G_\sigma(\mathrm{A}_{rgb})],
\end{equation}
where $m=1,2,\dots,M$; $\tau$ is the temperature hyperparameter of CLIP; $<\cdot,\cdot>$ denotes the calculation of cosine similarity; $G_\sigma$ represents Gaussian filter\cite{anomalyclip, zuma}; and $\operatorname{Bilinear}$ refers to bilinear interpolation. For the point cloud input, after deriving the multi-view anomaly score map $\mathrm{A}^{(m)}_{pc} \in \mathbb{R}^{H \times W}$ and anomaly response maps $\mathrm{S}^{(m)}_{pc} \in \mathbb{R}^{H \times W}$ in a similar manner from $\{\hat{\mathcal{L}}^{(m)}_k\}$, we adopt the post-processing method $\operatorname{BackTo3D}$ from \cite{pointad} to map $\mathrm{A}^{(m)}_{pc}$ and $\mathrm{S}^{(m)}_{pc}$ into a format corresponding to $\mathrm{S}_{rgb}$.
This mapping is defined as:
$
    \mathrm{A}_{pc}=\operatorname{BackTo3D}(\mathrm{A}^{(1)}_{pc},\mathrm{A}^{(2)}_{pc},\dots,\mathrm{A}^{(M)}_{pc}), \;
    \mathrm{S}_{pc}=\operatorname{BackTo3D}(\mathrm{S}^{(1)}_{pc},\mathrm{S}^{(2)}_{pc},\dots,\mathrm{S}^{(M)}_{pc}).
$
The resulting maps are denoted as $\mathrm{A}_{pc} \in \mathbb{R}^{H \times W}$ and $\mathrm{S}_{pc} \in \mathbb{R}^{H \times W}$. Although $\mathrm{S}_{rgb}$ and $\mathrm{S}_{pc}$ exhibit a strict correspondence, their reliability varies dynamically due to the inherent representational capacities of different modalities. Consequently, the final multimodal anomaly response map cannot be derived through simple averaging.

We utilize the hybrid loss $L_{hyb}$ proposed by PointAD\cite{pointad} to supervise the anomaly segmentation task. The definition of $L_{hyb}$ can be found in PointAD or the Supplement.


\subsection{Reliability-Aware Modality Response Balancing}

To explicitly model modality reliability, we propose Modality Reliability Calibration (MRC), a reliability-aware modality response balancing module. Given the point cloud response $\mathrm{S}_{pc}$ and RGB response $\mathrm{S}_{rgb}$, MRC module estimates the reliability of each modality and adaptively assigns fusion weights. Specifically, we adopt a simple yet effective cross-attention mechanism to achieve bidirectional feature interaction and reliability estimation across modalities. Initially,$\{\mathcal{L}^{(m)}_\mathrm{I}\}$ and $\{\hat{\mathcal{L}}^{(m)}_k\}$ are passed through $K_l$ cross-attention layers to enable deep feature fusion. Crucially, based on the observation that point clouds offer superior reliability compared to RGB images in zero-shot AD, we designate the point cloud features $\{\hat{\mathcal{L}}^{(m)}_k\}$ as the Key/ Value and RGB features as the Query. This strategic choice allows us to leverage geometric information to gauge the credibility of the RGB features, thereby producing the reliability map $\mathrm{W} \in \mathbb{R}^{H \times W}$ for the RGB modality. This procedure is mathematically defined as: 
\begin{equation}
    \mathcal{Z}_{i+1}=\operatorname{CrossAtt(\mathcal{Z}_i,\{\hat{\mathcal{L}}^{(m)}_k\})},i=0,1,\dots,K_l-1,
\end{equation}
\begin{equation}
    \mathrm{W}=\operatorname{FC}(\mathcal{Z}_{K_l}),
\end{equation}
where $\mathcal{Z}_0=\{\mathcal{L}^{(m)}_\mathrm{I}\}$ and $\operatorname{FC}$ denotes a $D$-dimensional single-layer perceptron. Ultimately, the multimodal anomaly score map $\mathrm{A}$ and the multimodal anomaly response map $\mathrm{S}$ are computed as:
\begin{equation}
    \mathrm{A}=\mathrm{W} \circ \mathrm{A}_{rgb} + (1-\mathrm{W}) \circ \mathrm{A}_{pc},
\end{equation}
\begin{equation}
    \mathrm{S}= \operatorname{Bilinear}[G_\sigma(\mathrm{A})],
\end{equation}
where $\circ$ denotes Hadamard product. We design a multimodal response loss $L_{mul}$ specifically for MRC as follows:
\begin{equation}
\begin{split}
    L_{mul}&=\operatorname{Focal}[\mathrm{A} \oplus (\mathbf{1}-\mathrm{A}),\mathrm{I}^{gt}]\\
    &+\operatorname{Dice}(\mathrm{A},\mathrm{I}^{gt})+\operatorname{Dice}(\mathbf{1}-\mathrm{A},\mathbf{1}-\mathrm{I}^{gt}),
\end{split}
\end{equation}
where $\operatorname{Focal}$ and $\operatorname{Dice}$ refer to the Focal Loss\cite{8237586} and Dice loss\cite{7785132}, respectively.


\subsection{Overall Loss\&Training}

The overall loss function is formulated as follows:
\begin{equation}
    L_{all}=\lambda_{fe} L_{fe}+\lambda_{hyb} L_{hyb}+\lambda_{mul} L_{mul},
\end{equation}
where $\lambda_{fe}$, $\lambda_{hyb}$, and $\lambda_{mul}$ are hyperparameters representing the weights of the respective loss terms. During training, we freeze the weights of both $\Psi_\mathrm{v}$ and $\Psi_\mathrm{t}$to fully leverage the robust generalization capabilities of CLIP. By minimizing the loss $L_{all}$, we jointly optimize MID module, MRC module, and the language vectors $\mathbf{P}_n$ and $\mathbf{P}_a$.


\section{Experiments}

\subsection{Experimental Setup}

\paragraph{Datasets.}

We evaluate our method on two widely adopted benchmark datasets for multimodal AD: MVTec 3D-AD\cite{m3dad} and Eyecandies\cite{eyec}. Detailed descriptions of these two datasets are provided in the supplement.


\paragraph{Evaluation metrics.} 
To characterize the performance limits of existing zero-shot approaches under practical conditions, we employ the stricter threshold-based metrics ${\mathrm{m}F_\mathrm{1}}^{.2}_{.8}$, $\mathrm{mAcc}^{.2}_{.8}$, and $\mathrm{mIoU}^{.2}_{.8}$\cite{metric}. These metrics evaluate anomaly localization over an industry-relevant confidence interval $[0.2, 0.8]$, jointly accounting for the precision–recall trade-off, overall classification accuracy, and spatial localization fidelity. In contrast to traditional metrics\cite{m3dad}—including image-level AUROC, pixel-level AUROC, AUPRO, and AP—that often exhibit saturation on contemporary zero-shot AD benchmarks\cite{metric}, these threshold-based metrics impose stronger penalties on abnormal responses in anomaly-free regions. Consequently, they offer superior discriminative power and better align with practical application scenarios. Additionally, we incorporate pixel-level AUROC (P-AUROC) and image-level AP as complementary metrics, with the corresponding results reported in the supplement.


\paragraph{Baselines.}
We compare WOOPS with representative zero-shot unimodal AD and zero-shot multimodal AD methods, including AnomalyCLIP\cite{anomalyclip}, FAPrompt\cite{faprompt}, GS-CLIP\cite{gs-clip}, PointAD\cite{pointad}, and ZUMA-FT\cite{zuma}. A brief description of the baselines is provided in the Supplement.

\paragraph{Implementation details.}
We utilize the pre-trained CLIP\cite{clip} (ViT-L/14@336px) as our backbone. Following the setup of prior methods\cite{pointad, zuma}, we set the number of multi-view projections to $M=9$, the feature dimension to $D=768$, and the length of learnable language vectors to $l=12$. To ensure a fair comparison with PointAD\cite{pointad}, we set the loss weight $\lambda_{hyb}=1.0$ of the corresponding loss term $L_{hyb}$. The remaining loss weights are configured as $\lambda_{fe}=1.0$ and $\lambda_{mul}=0.5$. We set the number of selected multi-view projections in the MID module to $X=1$. MRC module comprises $K_l=4$ cross-attention layers. We employ the Adam optimizer\cite{adam} with a learning rate of 0.001. The model is trained for $15$ epochs. All experiments are conducted on a single NVIDIA RTX 4090 GPU. A sensitivity analysis of the hyperparameters is provided in the supplementary material.


\begin{table*}[ht]
\centering
\begin{tabular}{cccc|ccc|ccc} 
\toprule
\multirow{2}{*}{Inference}  & \multirow{2}{*}{Training} & \multirow{2}{*}{Method} & \multirow{2}{*}{Source} & \multicolumn{3}{c|}{MVTec 3D-AD} & \multicolumn{3}{c}{Eyecandies}  \\ 
\cmidrule{5-10}
                     &                     &                         &                         & ${\mathrm{m}F_\mathrm{1}}^{.2}_{.8}$  & $\mathrm{mAcc}^{.2}_{.8}$ & $\mathrm{mIoU}^{.2}_{.8}$               & ${\mathrm{m}F_\mathrm{1}}^{.2}_{.8}$  & $\mathrm{mAcc}^{.2}_{.8}$ & $\mathrm{mIoU}^{.2}_{.8}$              \\ 
\midrule
\multirow{5}{*}{RGB} & RGB                 & AnomalyCLIP             & ICLR'24                 & 2.1  & 7.9  & 1.0                & 1.4  & 7.4  & 0.7               \\
                     & RGB                 & FAPrompt                & ICCV'25                 & 2.8  & 9.7  & 1.5                & 2.5  & 8.8  & 1.2               \\
                     & PC                  & PointAD (RGB)           & NIPS'24                 & 19.8 & 25.1 & 11.6               & \cellcolor{gray!10}13.1 & 18.0 & \cellcolor{gray!10}7.4               \\
                     & PC                  & ZUMA-FT (RGB)           & TPAMI'26                & \cellcolor{gray!10}20.1 & \cellcolor{gray!10}25.3 & \cellcolor{gray!10}11.8               & 13.0 & \cellcolor{gray!10}18.1 & 7.3               \\
                     & MUL                 & Ours (RGB)              & -                       & \cellcolor{gray!30}21.0 & \cellcolor{gray!30}44.7 & \cellcolor{gray!30}12.1               & \cellcolor{gray!30}17.2 & \cellcolor{gray!30}42.7 & \cellcolor{gray!30}10.3              \\ 
\midrule
\multirow{4}{*}{PC}  & PC                  & PointAD (PC)            & NIPS'24                 & 15.9 & \cellcolor{gray!10}29.6 & 9.5                & 14.7 & 28.1 & 8.4               \\
                     & PC                  & GS-CLIP                 & CVPR'26                 & \cellcolor{gray!10}17.6 & 24.8 & \cellcolor{gray!10}10.6               & \cellcolor{gray!10}16.1 & 24.6 & \cellcolor{gray!10}9.3               \\
                     & PC                  & ZUMA-FT (PC)            & TPAMI'26                & 16.9 & 28.6 & 10.1               & 14.8 & \cellcolor{gray!10}28.7 & 8.5               \\
                     & MUL                 & Ours (PC)               & -                       & \cellcolor{gray!30}22.6 & \cellcolor{gray!30}40.6 & \cellcolor{gray!30}13.8               & \cellcolor{gray!30}22.4 & \cellcolor{gray!30}32.9 & \cellcolor{gray!30}13.8              \\ 
\midrule
\multirow{3}{*}{MUL} & PC                  & PointAD                 & NIPS'24                 & \cellcolor{gray!30}20.6 & 33.1 & \cellcolor{gray!30}12.3               & \cellcolor{gray!10}15.8 & 28.1 & \cellcolor{gray!10}9.1               \\
                     & PC                  & ZUMA-FT                 & TPAMI'26                & 19.2 & \cellcolor{gray!30}57.9 & \cellcolor{gray!10}11.6               & 14.8 & \cellcolor{gray!10}43.8 & 8.6               \\
                     & MUL                 & Ours                    & -                       & \cellcolor{gray!10}19.3 & \cellcolor{gray!10}54.4 & \cellcolor{gray!10}11.6               & \cellcolor{gray!30}18.9 & \cellcolor{gray!30}45.0 & \cellcolor{gray!30}11.6              \\
\bottomrule
\end{tabular}
\caption{Comparison with state-of-the-art methods on MVTec 3D-AD dataset and Eyecandies dataset. Results are ranked separately according to the inference modality configurations. The \colorbox{gray!30}{best} and \colorbox{gray!10}{second-best} results within each group are denoted by gray shading.}
\label{tab:comp}
\end{table*}

\begin{figure}[t]
    \centering

    \begin{minipage}{0.05\linewidth}
        \centering
        \rotatebox{90}{\footnotesize Input}
    \end{minipage}%
    \begin{minipage}{0.9\linewidth}
        \centering
        \includegraphics[width=\linewidth]{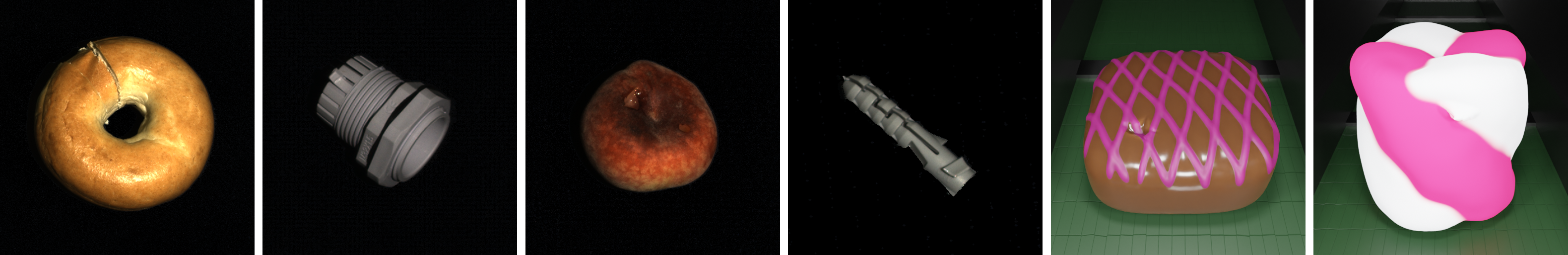}
    \end{minipage}

    \vspace{0.1mm}

    \begin{minipage}{0.05\linewidth}
        \centering
        \rotatebox{90}{\normalsize $\mathrm{A}_{rgb}$}
    \end{minipage}%
    \begin{minipage}{0.9\linewidth}
        \centering
        \includegraphics[width=\linewidth]{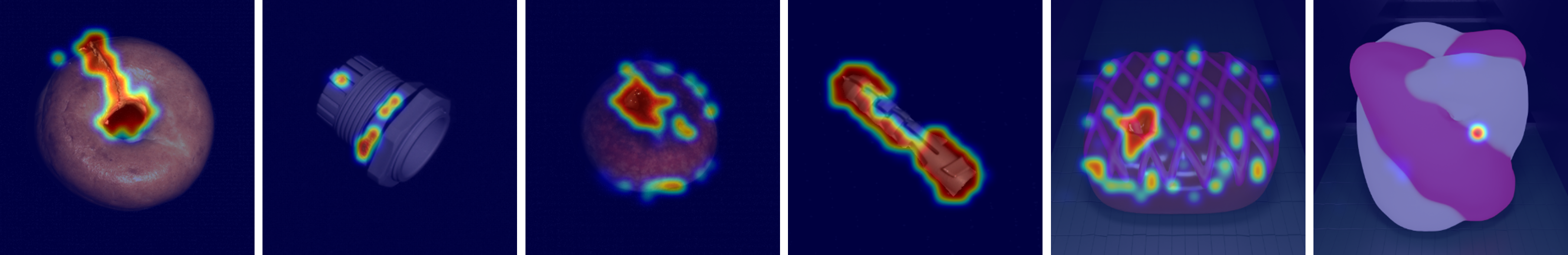}
    \end{minipage}

    \vspace{0.1mm}

    \begin{minipage}{0.05\linewidth}
        \centering
        \rotatebox{90}{\normalsize $\mathrm{W}$}
    \end{minipage}%
    \begin{minipage}{0.9\linewidth}
        \centering
        \includegraphics[width=\linewidth]{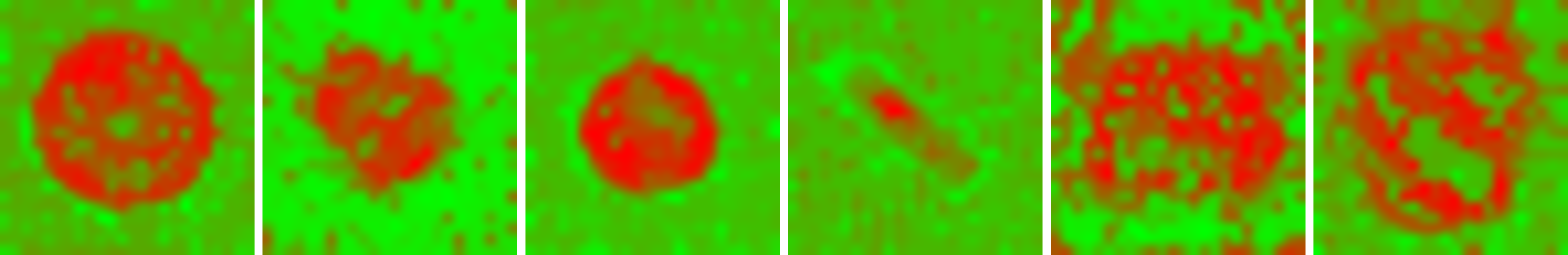}
    \end{minipage}

    \vspace{0.1mm}

    \begin{minipage}{0.05\linewidth}
        \centering
        \rotatebox{90}{\normalsize $\mathrm{A}_{pc}$}
    \end{minipage}%
    \begin{minipage}{0.9\linewidth}
        \centering
        \includegraphics[width=\linewidth]{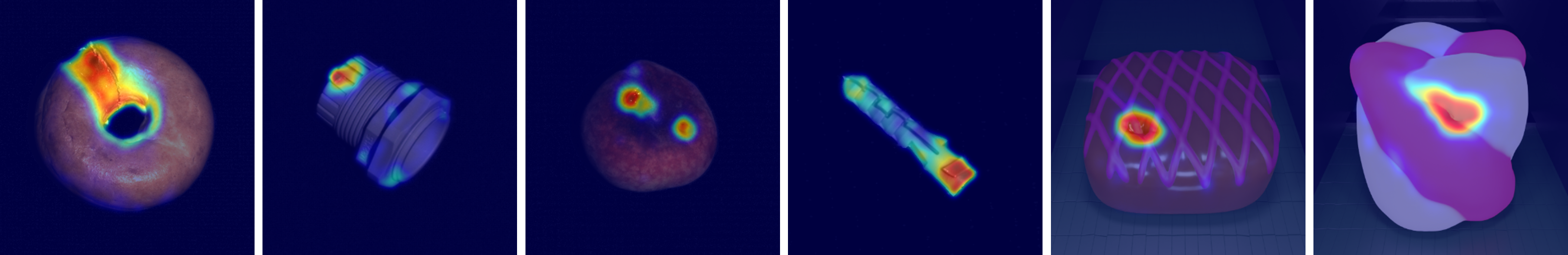}
    \end{minipage}

    \vspace{0.1mm}

    \begin{minipage}{0.05\linewidth}
        \centering
        \rotatebox{90}{\normalsize $\mathrm{A}$}
    \end{minipage}%
    \begin{minipage}{0.9\linewidth}
        \centering
        \includegraphics[width=\linewidth]{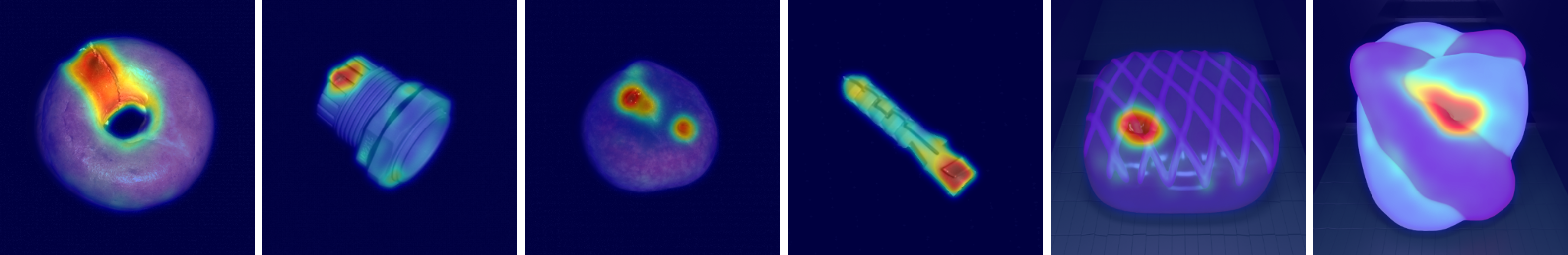}
    \end{minipage}

    \vspace{0.1mm}

    \begin{minipage}{0.05\linewidth}
        \centering
        \rotatebox{90}{\normalsize GT}
    \end{minipage}%
    \begin{minipage}{0.9\linewidth}
        \centering
        \includegraphics[width=\linewidth]{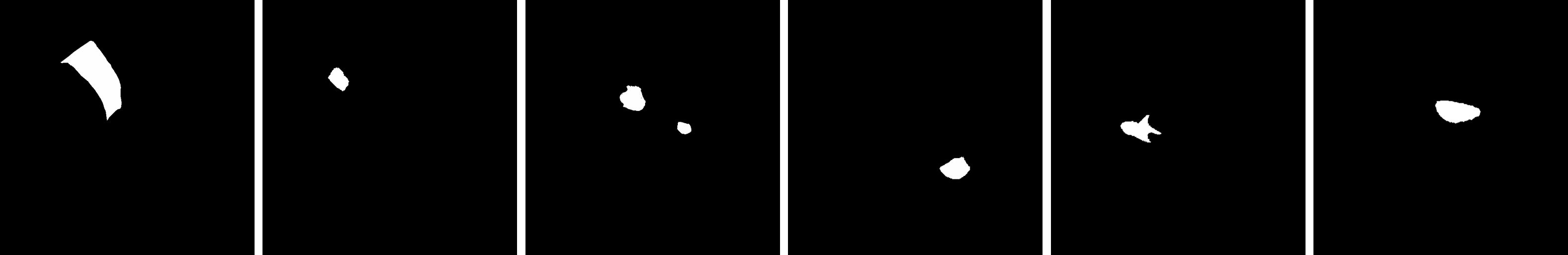}
    \end{minipage}

    \caption{Visualization results of WOOPS. From top to bottom: RGB input, RGB anomaly response, RGB reliability map, point cloud anomaly response, multimodal anomaly response, and ground truth. In the reliability map, warmer red tones indicate higher confidence, whereas cooler green tones denote lower reliability. As evident, regions exhibiting erroneous RGB responses correspond to low-reliability areas, demonstrating the effectiveness of our reliability calibration.}
    \label{fig:vis}
\end{figure}

\subsection{Main Results}

\paragraph{Comparison with state-of-the-art methods.}
We evaluated the performance of our proposed WOOPS against current state-of-the-art zero-shot AD methods on both the MVTec 3D-AD and Eyecandies datasets, with quantitative results summarized in Table \ref{tab:comp}. Our approach achieves the best or competitive performance across RGB-only, point cloud-only, and multimodal inference settings. Specifically, under point cloud-only inference, WOOPS substantially outperforms PointAD and ZUMA-FT, improving ${\mathrm{m}F_\mathrm{1}}^{.2}_{.8}$, ${\mathrm{mAcc}}^{.2}_{.8}$, and ${\mathrm{mIoU}}^{.2}_{.8}$ by 5.7\%, 11.0\%, and 3.2\%, respectively. This advantage stems from their lack of dedicated processing for multi-view projected images, which renders them vulnerable to interference from heterogeneous information. Although GS-CLIP incorporates depth maps and geometric cues to enhance prompt semantics, its performance still lags behind ours. In the multimodal setting, our method attains the best-reported results on the Eyecandies dataset, surpassing the previous best-performing approach by 3.1\%, 16.9\%, and 2.5\% in terms of ${\mathrm{m}F_\mathrm{1}}^{.2}_{.8}$, ${\mathrm{mAcc}}^{.2}_{.8}$, and ${\mathrm{mIoU}}^{.2}_{.8}$, respectively. On the MVTec 3D-AD dataset, WOOPS also demonstrates competitive performance, achieving an ${\mathrm{m}F_\mathrm{1}}^{.2}_{.8}$ of 19.3\% versus 20.6\%, ${\mathrm{mAcc}}^{.2}_{.8}$ of 54.4\% versus 57.9\%, and ${\mathrm{mIoU}}^{.2}_{.8}$ of 11.6\% versus 12.3\%. Results for RGB-only inference will be discussed in the following subsection. Qualitative comparisons and quantitative evaluations based on conventional metrics are provided in the Supplement.


\paragraph{Point clouds matter in zero-shot multimodal anomaly detection. }
\label{subsec:rgb}
We further evaluate RGB-only inference under different training modalities. As shown in Table \ref{tab:comp}, methods trained with RGB alone perform significantly worse than methods trained with point cloud or RGB-point cloud data. This phenomenon indicates that point cloud information facilitates the learning of discriminative semantic representations for normal and anomalous patterns, and benefits cross-category generalization beyond the point cloud modality itself. Since some point-cloud-based methods use vision-language models, they can still be applied to RGB-only inference. The superior RGB-only performance of our method indicates that the proposed point cloud feature enhancement also improves modality-transferable anomaly cues. This result further supports our central observation that point cloud data play a crucial role in zero-shot multimodal AD. Further visualization results in Fig. \ref{fig:vis} highlight the distinct behaviors of different modalities. While point cloud responses prove more reliable than their RGB counterparts, the fused multimodal response maintains high fidelity. Analysis of the RGB reliability map confirms that this reliability is achieved by mitigating spurious RGB responses within normal regions.


\paragraph{Analysis of the performance gap between point-cloud-only and multimodal settings. }
A notable observation is that the point-cloud-only version of our method performs better than its multimodal counterpart. This differs from the common expectation that multimodal inference should always outperform single-modality inference. The experimental analyses in the preceding two paragraphs demonstrate that, compared with point clouds, RGB inputs are more prone to spurious anomaly activations—a conclusion further corroborated by the qualitative results presented in Fig. \ref{fig:vis}. When the responses derived from point clouds exhibit low reliability, incorporating RGB responses as complementary information improves performance; however, when point cloud responses become sufficiently reliable due to the MID module, RGB responses instead exert a detrimental effect on performance. To the best of our knowledge, this work is the first to reveal the potential multimodal collapse in zero-shot multimodal AD, and we intend to investigate more principled and effective multimodal fusion strategies in future research.


\begin{table}[t]
\centering
\begin{tabular}{c|ccc|ccc} 
\toprule
\multirow{3}{*}{Setting} & \multicolumn{6}{c}{${\mathrm{m}F_\mathrm{1}}^{.2}_{.8}$}                                            \\ 
\cmidrule{2-7}
                         & \multicolumn{3}{c|}{MVTec 3D-AD} & \multicolumn{3}{c}{Eyecandies}  \\ 
\cmidrule{2-7}
                         & RGB  & PC   & MUL                & RGB  & PC   & MUL               \\ 
\midrule
\cellcolor{gray!20}Full                     & \cellcolor{gray!20}\textbf{21.0} & \cellcolor{gray!20}\textbf{22.6} & \cellcolor{gray!20}\textbf{19.3}               & \cellcolor{gray!20}\textbf{17.2} & \cellcolor{gray!20}\textbf{22.4} & \cellcolor{gray!20}\textbf{18.9}              \\
w/o MID                  & 18.5 & 17.4 & 18.1               & 13.1 & 13.8 & 13.3              \\
w/o MRC                  & 20.4 & 22.5 & 19.1               & 9.3  & 11.6 & 11.4              \\
w/o $L_{fe}$                  & 18.0 & 21.0 & 18.0               & 17.0 & 21.9 & 18.7              \\
w/o $L_{mul}$                 & 12.5 & 15.8 & 14.0               & 13.1 & 14.8 & 13.1              \\
w/o $L_{hyb}$                 & 1.5  & 4.8  & 1.5                & 1.2  & 2.6  & 2.6               \\
\bottomrule
\end{tabular}
\caption{Quantitative results of the ablation study. Top results are in bold, and the default setting is indicated by gray shading.}
\label{tab:ablation}
\end{table}

\begin{figure}[th]
    \centering
    \begin{subfigure}[b]{0.48\linewidth}
        \centering
        \includegraphics[width=\textwidth]{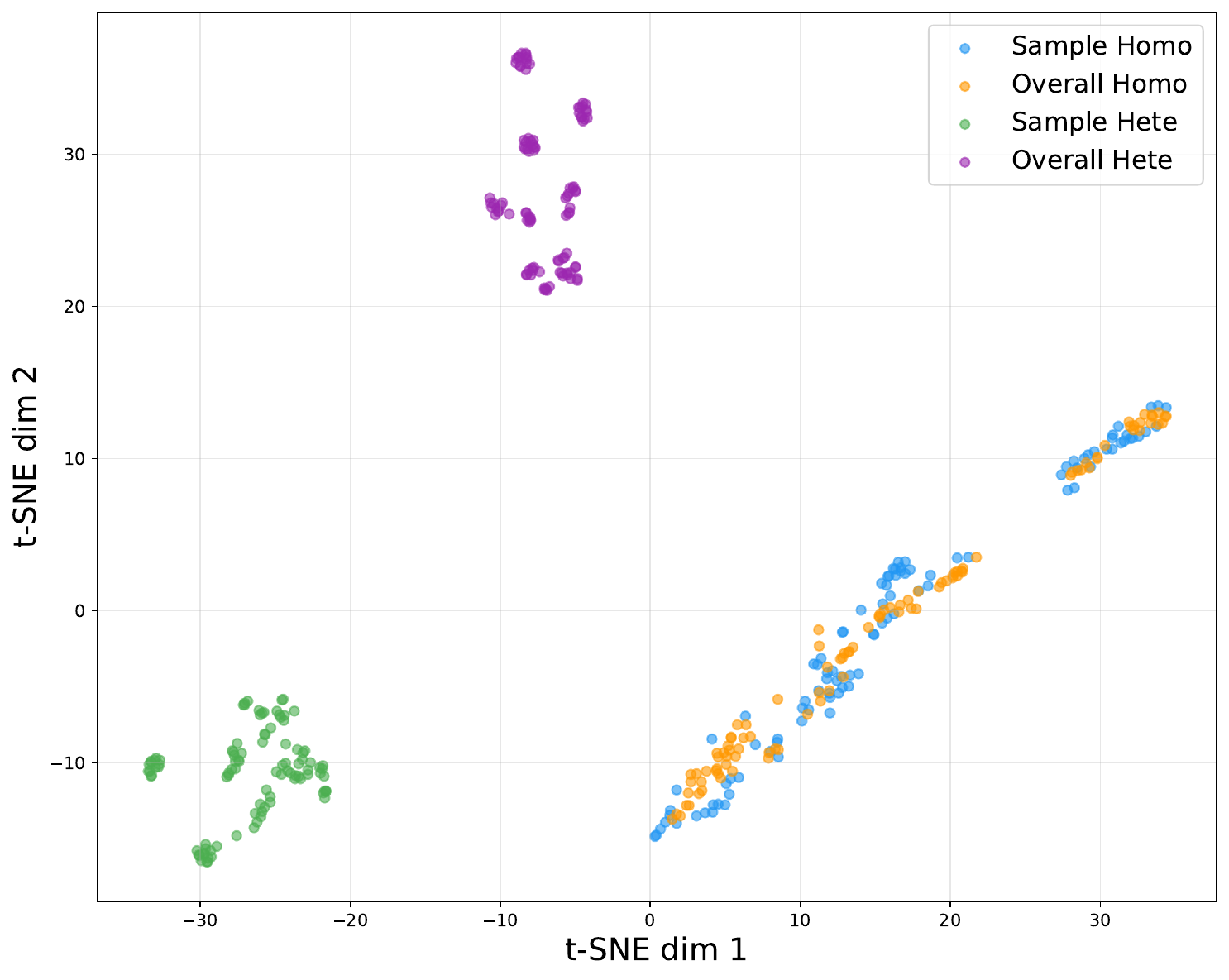}  
        \caption{Visualization of Global Feature Distributions via t-SNE}
    \end{subfigure}
    \hfill
    \begin{subfigure}[b]{0.48\linewidth}
        \centering
        \includegraphics[width=\textwidth]{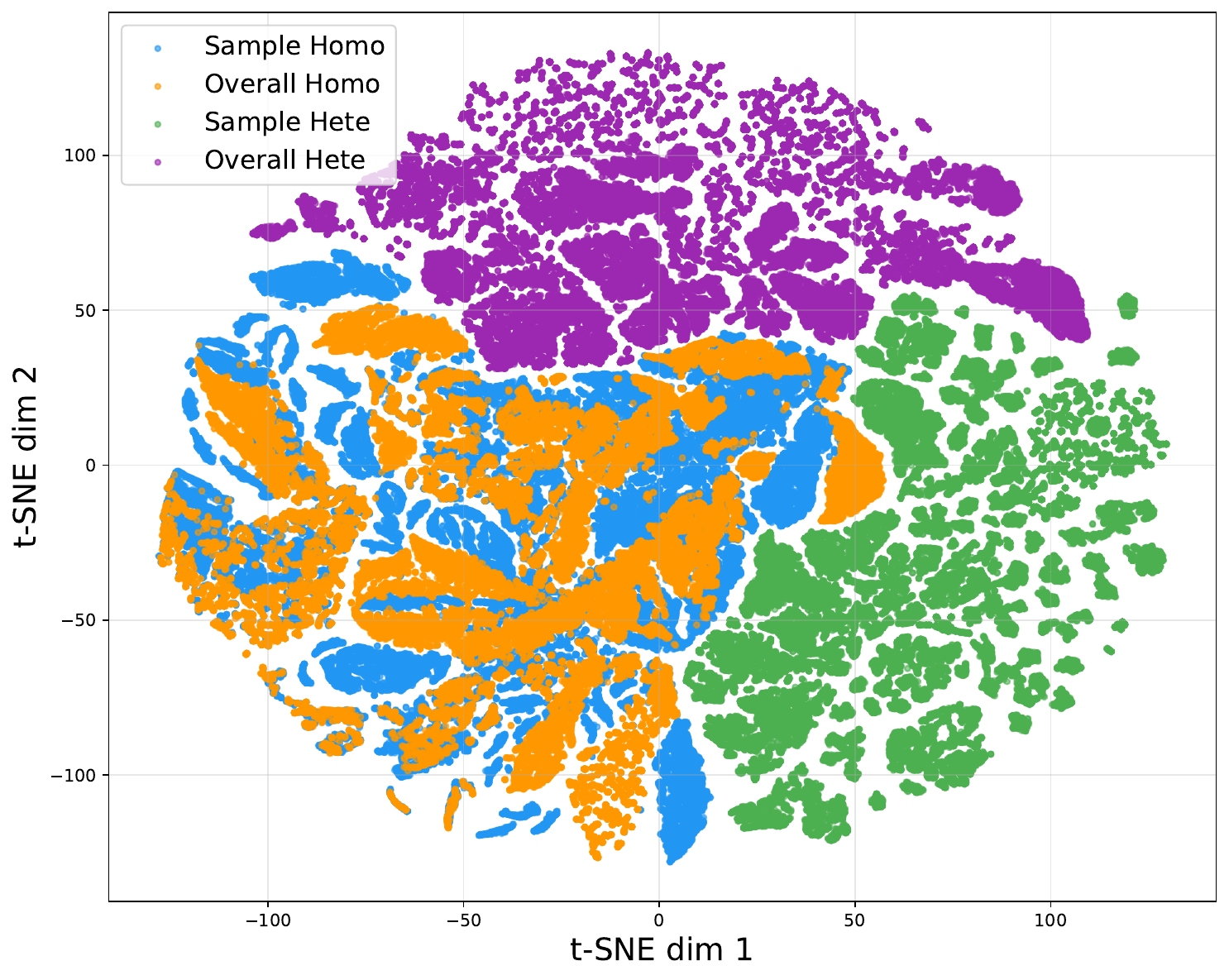}  
        \caption{Visualization of Local Patch Feature Distributions via t-SNE}
    \end{subfigure}
    
    \caption{t-SNE Visualization of Decoupled Features from the MID Module. "Sample" denotes the features of the projection views sampled by the MID module, while "Overall" refers to the point cloud features. Zoom in for details.}
    \label{fig:tsne}
\end{figure}

\subsection{Ablation Studies}

\paragraph{Effect of the MID module.}
We first ablate the proposed point cloud feature enhancement module. As shown in Table \ref{tab:ablation}, removing MID module leads to a clear performance drop, especially under point-cloud-only inference. This confirms that directly aggregating multi-view projected features is suboptimal. By suppressing unreliable view-specific features and enhancing consistent geometric cues, MID module produces higher-quality point cloud representations. We visualize the decoupled features from the MID module using t-SNE, as shown in Fig. \ref{fig:tsne}. As illustrated, driven by the decoupling loss $L_{dec}$, the homogeneous and heterogeneous information within both global and local patch features are effectively repelled. Furthermore, owing to the explicit constraints imposed by $L_{dc}$ and $L_{gmmd}$, the homogeneous components of the global features achieve precise alignment. Building upon this aligned global representation, and leveraging the powerful semantic priors embedded in the frozen CLIP visual encoder, the homogeneous information in local patch features is also naturally aligned.


\paragraph{Effect of the MRC module.}
We then evaluate the modality response balancing module. Table \ref{tab:ablation} shows that removing MRC module causes unstable performance across datasets. In particular, the model may perform well on one dataset but degrade substantially on another dataset across multiple modality settings. This indicates that fixed or uncalibrated fusion is sensitive to dataset-specific modality reliability. By adaptively calibrating modality contributions according to their reliability, MRC module improves robustness and reduces the risk of being dominated by unreliable modality responses.

\paragraph{Ablation Study of Loss Terms. }
To validate the effectiveness of the proposed loss terms, i.e., $L_{fe}$ and $L_{mul}$, we conduct ablation studies on these components, with quantitative results summarized in Table \ref{tab:ablation}. Removing $L_{fe}$ leads to a moderate performance degradation, confirming its necessity. Notably, in the absence of $L_{fe}$, the $\Phi _o$ acts as a post-hoc feature adapter\cite{Gao_2023}, thereby preventing a catastrophic drop in performance. In contrast, discarding $L_{mul}$ results in a significant decline, demonstrating that it serves as a crucial supervisory signal within our training framework. Furthermore, we supplement the ablation study with an analysis of $L_{hyb}$. The results indicate that removing $L_{hyb}$ causes a substantial performance drop, underscoring its role as a fundamental objective for multimodal zero-shot AD.


\section{Conclusion}

We revisit zero-shot multimodal AD by questioning a common assumption: RGB and point cloud modalities are equally reliable and should contribute equally. Under recently proposed stringent metrics, we show that point clouds provide more reliable cross-category generalization, while RGB may introduce false anomaly responses in normal regions. Guided by this finding, we propose WOOPS, which enhances point cloud representations through MID module and calibrates modality contributions through MRC module. Extensive experiments demonstrate the effectiveness of both geometry and reliability-aware modality utilization under zero-shot multimodal AD.

\bibliography{aaai2027}


\end{document}